%% file: technical-report.tex
\documentclass[10pt,a4paper]{article}

\usepackage[margin=19mm,top=17mm,bottom=20mm]{geometry}
\usepackage[T1]{fontenc}
\usepackage{lmodern}
\usepackage{microtype}
\usepackage{amsmath}
\usepackage{booktabs}
\usepackage{tabularx}
\usepackage{graphicx}
\usepackage{float}
\usepackage{placeins}
\usepackage{xcolor}
\usepackage{tikz}
\usetikzlibrary{arrows.meta,positioning}
\usepackage{enumitem}
\usepackage[font=small,labelfont=bf]{caption}
\usepackage{titlesec}
\usepackage[hidelinks]{hyperref}

\definecolor{reportteal}{HTML}{087F8C}
\definecolor{reportblue}{HTML}{286F9B}
\definecolor{reportorange}{HTML}{D97706}
\hypersetup{colorlinks=true,linkcolor=reportteal,citecolor=reportteal,urlcolor=reportblue}

\titleformat{\section}[block]{\large\bfseries}{\thesection}{0.55em}{}
\titleformat{\subsection}[block]{\normalsize\bfseries}{\thesubsection}{0.5em}{}
\titlespacing*{\section}{0pt}{1.0ex plus 0.4ex}{0.45ex}
\titlespacing*{\subsection}{0pt}{0.75ex plus 0.3ex}{0.3ex}
\setlist{nosep,leftmargin=1.4em}
\newcommand{\LaneGroup}{\texttt{LaneGroup}}
\newcommand{\reportcontentwidth}{0.82\linewidth}

\title{\vspace{-1.8em}\textbf{A Graph-Based Control Interface for Traffic Signals\\on Heterogeneous Road Networks}}
\author{Bertil Braun\\{\small\href{mailto:contact@bertil-braun.de}{contact@bertil-braun.de}}}
\date{}

\begin{document}
\maketitle
\vspace{-2.2em}

\begin{abstract}
We present a traffic-signal control interface in which a shared graph neural network assigns scores to individual traffic movements.
Each junction converts these scores into its own variable-sized set of legal signal phases using a deterministic incidence matrix.
Directed corridor nodes provide traffic context, while movement nodes represent controlled input-to-output paths through junctions.
Typed mean aggregation produces one scalar per movement; phase definitions and signal timing remain outside the learned network.
This makes graph size and junction-specific action count independent of the learned parameter shapes.
PPO experiments evaluate the interface on unseen synthetic grid geometries, altered signal coverage, and five heterogeneous city graphs.
The policies retained performance across unseen geometries within the synthetic grid family, while changes in signal coverage exposed sensitivity to a signal-coverage distribution shift.
A single trained city-policy instance executed across all five city graphs, with heterogeneous outcomes.
These results provide feasibility evidence rather than a general estimate of transfer to arbitrary road networks.
\end{abstract}
\small

\section{Introduction}

Traffic-signal action spaces are local.
A three-arm junction, a regular four-arm junction, and a junction with protected turns need not have the same number or meaning of phases.
Consequently, a fixed output called ``phase 2'' does not have reusable semantics across road networks.
Padding such a head changes its tensor shape but does not give the padded phase indices shared meaning across junctions.

We present a traffic-signal control interface in which a shared graph neural network assigns scores to individual traffic movements.
Each junction converts these scores into its own variable-sized set of legal signal phases using a deterministic incidence matrix.
The central design choice is therefore a narrow boundary: learning prioritizes comparable movement objects, while deterministic code constructs and operates each junction's valid local choices.

The resulting system combines a reusable movement-level graph representation, deterministic construction of junction-specific action spaces, and a feasibility evaluation across synthetic and city simulations.
Its variable graph and action dimensions are a structural property of the design.
We distinguish a structural property established by construction, implementation validation over heterogeneous graphs, and bounded empirical evidence about learned performance.
Accordingly, variable graph and action dimensions are the primary result, trainability and cross-scenario execution are secondary validation, and learned--baseline outcomes are exploratory comparisons under unmatched controller timings.
The empirical evaluation is organized around three research questions:
\begin{description}[leftmargin=4.2em,style=nextline]
  \item[RQ1 --- Transfer within the synthetic generator family.] How does the trained policy behave on unseen grid sizes and aspect ratios produced by the same generator?
  \item[RQ2 --- Distribution shift.] What happens when signal coverage changes?
  \item[RQ3 --- City feasibility.] Can the same trained policy execute on several heterogeneous city simulations, and how variable are the observed outcomes?
\end{description}

Movement pressure and movement-structured control have substantial prior lineage.
Max-pressure methods score compatible stages from constituent flows \cite{varaiya2013maxpressure}, while PressLight connects pressure to learned control \cite{wei2019presslight}.
FRAP learns phase demand and pairwise phase competition from movement features, explicitly structuring the learned phase values around conflict and symmetry \cite{zheng2019frap}.
TransferLight is the closest architectural comparison: its learned directed hierarchy aggregates lane segments into movement representations, movements into phase representations, and interactions among phases before producing phase energies \cite{schmidt2024transferlight}.
It also uses weight-tied decentralized agents, domain-randomized training networks, and learned movement-to-phase semantics.
In contrast, our learned actor operates on a city-level LaneGroup/Movement context graph and stops at one scalar per movement; phase membership, phase enumeration, incidence summation, and timing remain deterministic.
The narrower contribution is this transparent boundary between shared movement scoring and automatically constructed local action spaces, rather than movement-based control itself or a broader transfer claim.

\textbf{Scope.}
This report evaluates an implementation and architectural interface rather than proposing a new reinforcement-learning algorithm.
The experiments evaluate transfer within the specified simulation families and execution across heterogeneous action spaces.
They do not establish general transfer across arbitrary road networks.

\section{Movement-level control interface}

\subsection{Intuition and control objects}

A \emph{movement} is one legal controlled path from an incoming road corridor to an outgoing corridor; straight travel is a movement as well as a turn.
A \emph{phase} is a compatible set of movements that may receive green together.
The controller chooses one phase per junction, rather than setting individual lamps independently.

\begin{figure}[H]
  \centering
  \resizebox{0.73\linewidth}{!}{\input{figures/intersection-concepts.tex}}
  \caption{The control vocabulary. The shared model scores movements individually; each junction supplies its own compatible phase sets. The phase shown is illustrative, not a universal template.}
  \label{fig:concepts}
\end{figure}
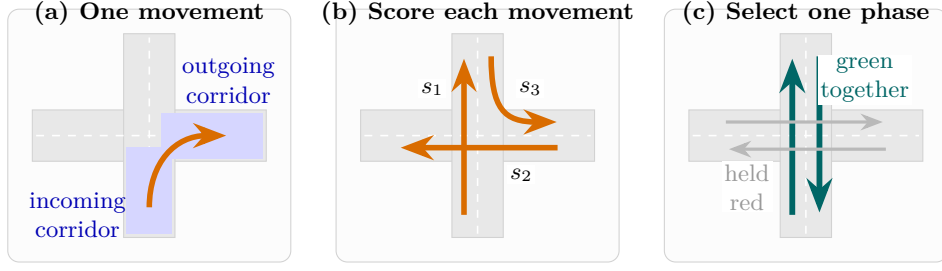
\FloatBarrier

The implementation groups consecutive directed road segments into a \LaneGroup{} when an unsignalized continuation is unambiguous.
Opposite directions remain separate because their queues, speeds, and destinations differ.
At a controlled junction, a legal input-to-output connection becomes a movement node.
The junction owns its movements and phases but is not itself a GNN node.

\begin{table}[H]
  \small
  \centering
  \caption{Roles in the representation and action interface.}
  \begin{tabularx}{\reportcontentwidth}{@{}lXX@{}}
    \toprule
    Object & Represents & Used for\\
    \midrule
    lane group & One direction of a road corridor & Queue, speed, occupancy, capacity, arrivals, departures, and downstream space\\
    movement & One controlled input-to-output path & Turn-local demand, controlled-link count, and recent green state\\
    phase & One compatible local movement set & A junction-specific selectable action\\
    \bottomrule
  \end{tabularx}
  \label{tab:terminology}
\end{table}

At each decision, features are refreshed from SUMO, the shared GNN emits one score per movement, and a junction-local incidence matrix converts those scores to phase logits.
An availability mask enforces minimum green; the runtime inserts yellow when an accepted target changes.
Only movement scoring is learned.

\subsection{Typed graph and message-passing architecture}

Each movement \(m\) has one input LaneGroup \(i(m)\) and one output LaneGroup \(o(m)\), giving four directed information relations:
\[
L_{\mathrm{in}}\!\rightarrow M,\quad
L_{\mathrm{out}}\!\rightarrow M,\quad
M\!\rightarrow L_{\mathrm{in}},\quad
M\!\rightarrow L_{\mathrm{out}}.
\]
The output-to-movement relation returns downstream storage context to a movement that would feed that corridor.
Unsignalized pass-through connections are weighted \(L\!\rightarrow L\) edges with \(w_{ql}=\exp(-t_{\mathrm{ff}}/30\,\mathrm{s})\).

Let \(\rho\) denote ReLU, \(\Vert\) concatenation, and \(x_l,x_m\) the LaneGroup and movement feature vectors.
The implementation first creates
\begin{align}
h_l^{(0)} &= \rho(E_Lx_l),\\
h_m^{(0)} &= \rho\!\left(E_M[x_m\Vert h_{i(m)}^{(0)}\Vert h_{o(m)}^{(0)}]\right).
\end{align}
For relation \(r\), target \(v\), and source embeddings \(z_q\), define the transformed mean
\[
\mathcal A_r^{(k)}(v;z)=
\frac{1}{\max(1,|\mathcal N_r(v)|)}
\sum_{q\in\mathcal N_r(v)} w_{qv}W_r^{(k)}z_q,
\]
where \(w_{qv}=1\) except on unsignalized connector edges.
The incoming and outgoing messages to movement \(m\) are
\begin{align}
a_{\mathrm{in},m}^{(k)}&=\mathcal A_{L_{\mathrm{in}}\to M}^{(k)}(m;h_L^{(k)}),&
a_{\mathrm{out},m}^{(k)}&=\mathcal A_{L_{\mathrm{out}}\to M}^{(k)}(m;h_L^{(k)}).
\end{align}
One block first updates the movement embedding:
\begin{align}
h_m^{(k+1)}=\rho\!\left(U_M^{(k)}
[h_m^{(k)}\Vert a_{\mathrm{in},m}^{(k)}\Vert a_{\mathrm{out},m}^{(k)}]\right).
\label{eq:movement-update}
\end{align}
It then forms the return messages
\begin{align}
b_{\mathrm{in},l}^{(k)}&=\mathcal A_{M\to L_{\mathrm{in}}}^{(k)}(l;h_M^{(k+1)}),&
b_{\mathrm{out},l}^{(k)}&=\mathcal A_{M\to L_{\mathrm{out}}}^{(k)}(l;h_M^{(k+1)}),\\
c_l^{(k)}&=\mathcal A_{L\to L}^{(k)}(l;h_L^{(k)}),
\end{align}
and updates the LaneGroup embedding:
\begin{align}
h_l^{(k+1)}=\rho\!\left(U_L^{(k)}
[h_l^{(k)}\Vert(b_{\mathrm{in},l}^{(k)}+c_l^{(k)})\Vert b_{\mathrm{out},l}^{(k)}]\right).
\label{eq:lane-update}
\end{align}
Every relation has its own linear map, aggregation is a mean rather than attention, and return messages use the just-updated embeddings from \eqref{eq:movement-update}.
An empty neighbourhood produces the zero vector.
After two blocks, an MLP maps every \(h_m^{(2)}\) to one scalar \(s_m\).
Each junction is treated as a parameter-sharing local agent: the critic mean-pools that junction's movement embeddings, returns one value, and its local reward and value sequence produce a separate temporal GAE stream.
For PPO updates, variable-size state graphs are concatenated as disconnected packed graphs; junctions with matching local movement and phase dimensions are grouped for batched value and incidence operations, and minibatches are budgeted by junction/action samples rather than padded to a universal graph or action size.

\begin{figure}[H]
  \centering
  \resizebox{0.64\linewidth}{!}{\input{figures/representation.tex}}
  \caption{The typed representation and update order. With this order, two blocks allow information from one movement to reach another through a shared LaneGroup.}
  \label{fig:representation}
\end{figure}
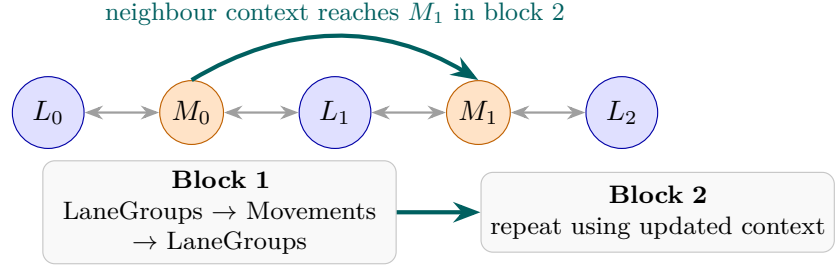
\FloatBarrier

The parameter matrices above are shared over all nodes and edges of a type.
Their shapes depend only on feature and hidden dimensions.
The parameter shapes are therefore independent of graph and action-space size.

\subsection{Deterministic local action spaces}

The build pipeline obtains controlled links and request-conflict data from SUMO \texttt{netconvert} \cite{alvarezlopez2018sumo}.
Links constrained to activate together form atomic groups; groups with internal SUMO conflicts are rejected.
Two groups are incompatible when SUMO reports a conflict in either direction or different incoming approaches merge into the same outgoing edge.
Bron--Kerbosch enumeration \cite{bron1973cliques} produces all maximal compatible group sets, each of which becomes a selectable phase.
Here, maximal means that no additional compatible group can be added, not that the phase has maximum size.

For junction \(j\), \(A_j\in\{0,1\}^{|P_j|\times|M_j|}\) records whether phase \(p\) enables movement \(m\).
The local logits are
\[
\boldsymbol\ell_j=A_j\mathbf s_j,
\qquad
\ell_{j,p}=\sum_{m\in M_j}A_{j,p,m}s_m.
\]
For example,
\[
A_j=
\begin{bmatrix}1&1&0&0\\0&0&1&1\end{bmatrix},\qquad
\mathbf s_j=
\begin{bmatrix}1.2&0.7&0.6&-0.4\end{bmatrix}^{\mathsf T},\qquad
A_j\mathbf s_j=
\begin{bmatrix}1.9&0.2\end{bmatrix}^{\mathsf T}.
\]
The resulting logits favor the first phase; masking and categorical sampling follow before the runtime applies any signal transition.
A different junction may provide a \(6\times11\) matrix without changing the scorer.

\begin{figure}[H]
  \centering
  \resizebox{0.78\linewidth}{!}{\input{figures/phase-interface.tex}}
  \caption{Offline phase construction and online action selection. The fixed incidence matrix changes with each junction; the learned scorer does not.}
  \label{fig:phase-interface}
\end{figure}
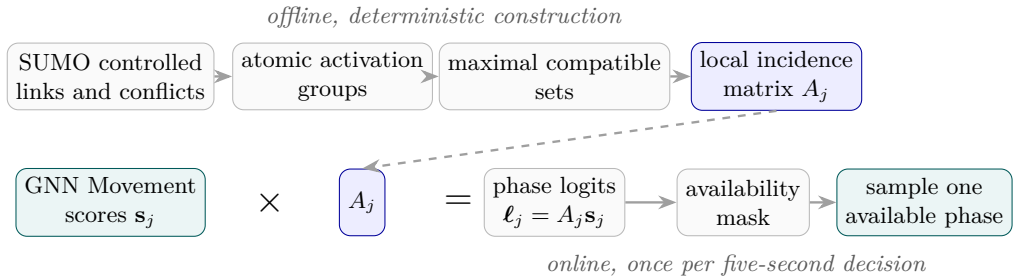
\FloatBarrier

Sum aggregation intentionally restricts the policy class: a phase logit is the additive utility of its enabled movements, so the actor cannot directly represent arbitrary within-phase interactions.
Phases sharing movements also have correlated logits, and when scores are positive a larger phase can receive a higher logit simply because it has more terms.
The restriction keeps the learned/local interface transparent, but neither its phase-size effect nor alternatives such as normalized or learned phase aggregation were evaluated separately.

\section{Experimental design and reproducibility}

\subsection{Training protocol and local reward}

PPO \cite{schulman2017ppo} optimizes the complete policy.
All reported studies use two message-passing blocks, \(5\,\mathrm{s}\) decisions, immediate \(3\,\mathrm{s}\) yellow on a change, one decision of minimum green, four PPO epochs, 200 decisions per rollout, and entropy coefficient \(0.001\).
Rollout counts are allocated to approximately balance junction/action samples across differently sized training graphs.

The code assigns one reward to each junction \(j\) per decision interval \(\Delta=5\,\mathrm{s}\).
Qualitatively, the local reward encourages vehicles to move through and leave the incoming approaches while penalizing deceleration and queues.
Let \(I_j\) be its unique incoming lanes, \(D_j=\sum_{l\in I_j}D_l\) their total length, \(n_l\) the vehicle count, \(\bar v_l\) mean speed, \(v_l^{\max}\) speed limit, and \(V_j(t)\) the set of vehicle IDs on those lanes.
With \(f_l(t)=\operatorname{clip}(\bar v_l(t)/v_l^{\max},0,1)\), the implemented terms are:

\begin{table}[H]
  \footnotesize
  \centering
  \caption{Exact local reward terms for the reported runs. Values are sampled at decision boundaries.}
  \begin{tabularx}{\reportcontentwidth}{@{}lXl@{}}
    \toprule
    Term & Definition and interpretation & Unit\\
    \midrule
    Progress \(p_j\) & \(D_j^{-1}\sum_{l\in I_j}n_l(t)f_l(t)\): speed-normalized moving-vehicle density & veh/m\\
    Discharge \(d_j\) & \(|V_j(t-\Delta)\setminus V_j(t)|/(\Delta D_j)\): IDs leaving the incoming-lane set & veh/(m\,s)\\
    Braking \(b_j\) & \((\Delta D_j)^{-1}\sum_{l\in I_j}n_l(t)[\bar v_l(t-\Delta)-\bar v_l(t)]_+/v_l^{\max}\) & veh/(m\,s)\\
    Gridlock \(q_j\) & \(D_j^{-1}\sum_{l\in I_j}n_l(t)(1-f_l(t))\): speed-deficit density & veh/m\\
    \bottomrule
  \end{tabularx}
  \label{tab:reward}
\end{table}

The first braking sample is zero because no preceding snapshot exists.
The per-junction reward is
\[
r_j=\operatorname{clip}_{[-1,1]}\!\left(p_j+10d_j-10b_j-0.02q_j\right).
\]
Global delay, completed-network flow, direct throughput, phase switching, and teleport terms have zero weight in these runs.
Because the terms have different units and are manually weighted, the final reward is a dimensionless optimization objective.
It is a local surrogate, not a guarantee of improvement in network-level throughput, completion, or wait density: for example, discharge from one junction can move vehicles into a congested downstream region.

\subsection{Evaluation metrics}

Every episode records per-seed metrics before arithmetic means are formed.
For an episode of \(T\) simulated seconds, let \(C\) be trips with a SUMO \texttt{tripinfo} arrival after warm-up and \(D\) the initial post-warm-up population plus subsequent departures.

\begin{table}[H]
  \footnotesize
  \centering
  \caption{Primary evaluation metrics and aggregation.}
  \begin{tabularx}{\reportcontentwidth}{@{}lXl@{}}
    \toprule
    Metric & Implemented definition & Unit\\
    \midrule
    Throughput & \(C/(T/3600)\) & veh/h\\
    Completion & \(C/D\); vehicles still active at the horizon remain incomplete & fraction or \%\\
    Wait density & Time mean of \(\sum_l W_l(t)/\sum_l D_l\), where \(W_l\) is SUMO's total waiting time of vehicles currently on lane \(l\) & s/m\\
    \bottomrule
  \end{tabularx}
  \label{tab:metrics}
\end{table}

Wait density uses all non-internal network lanes at every simulated second and exposes queues left at the horizon.
Grid results separate three policy-training seeds from six held-out traffic seeds.

\subsection{Baseline protocol and study matrix}

All policies receive the same saved SUMO network, routes, initial-occupancy range, warm-up, episode seed, conflict-derived phase set, minimum-green rule, and yellow transition.
The learned and uniform-random policies choose every \(5\,\mathrm{s}\); fixed time cycles through phase order every \(10\,\mathrm{s}\); max pressure and queue recompute a target every \(10\,\mathrm{s}\).
This mismatch is a timing confound: the comparisons evaluate the recorded controller implementations and do not isolate learning or policy architecture from control frequency.
Max pressure observes the instantaneous halting counts in the downstream detector regions of the movement's input and output LaneGroups and scores \(n^{\mathrm{halt}}_{\mathrm{in}}-n^{\mathrm{halt}}_{\mathrm{out}}\); queue observes the input detector count only.
Both sum movement scores through the same \(A_j\) and retain the current phase on a tie.
Fixed time and uniform random use no traffic observation.
The \(10\,\mathrm{s}\) durations and all other baseline parameters were fixed in the experiment configurations; the records contain no baseline-tuning study.
Max pressure is the reference for RQ1 and RQ2.
For RQ3, city-specific comparisons also report the highest-throughput non-learned baseline: max pressure in Karlsruhe, queue in Mannheim, and fixed time in Stuttgart, Heidelberg, and Freiburg.

\begin{itemize}
  \item \textbf{RQ1:} three mixed-grid training seeds, trained on five square or rectangular geometries whose longest axis is five; final evaluation on ten geometries at demands \(0.6,0.7,0.8\), including \(6\times6\), with six held-out traffic seeds.
  \item \textbf{RQ2:} a full-coverage-trained \(6\times6\) policy evaluated after removing signals, plus one separate training seed exposed to five \(50\%\)-coverage \(4\times4\) layouts and evaluated on held-out \(25\%\)--\(100\%\) layouts.
  \item \textbf{RQ3:} one city policy-training seed with rollouts from Karlsruhe, Mannheim, Heidelberg, and Freiburg. Stuttgart contributed no training rollouts. Although periodic Stuttgart evaluations were available, the reported iteration-60 checkpoint was the fixed terminal checkpoint rather than one selected using Stuttgart performance. It was evaluated for \(1200\,\mathrm{s}\) at demand scale 1.0 on three traffic seeds in all five cities.
\end{itemize}
All reported comparisons use libsumo/sumolib 1.27.1 as resolved by \texttt{uv.lock}.

Sampled execution is primary because it corresponds to the stochastic policy optimized during PPO training.
Greedy execution performed worse in the city evaluation.
The sampled results therefore do not imply equivalent performance under deterministic argmax execution.
Deployment under deterministic action selection would require separate validation.

\section{Structural property and empirical results}

\begin{figure}[H]
  \centering
  \includegraphics[width=0.88\linewidth]{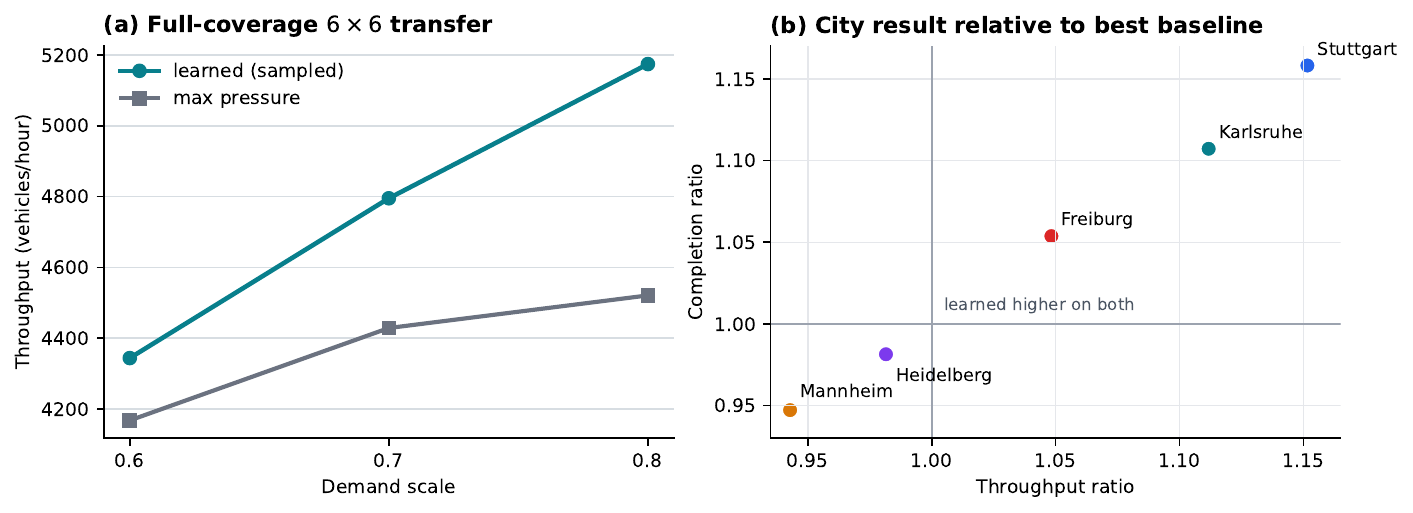}
  \caption{Observed feasibility results. Left: sampled \(6\times6\) performance by demand. Right: iteration-60 city outcomes; each city is normalized by its own highest-throughput non-learned baseline, so baseline identities differ across cities.}
  \label{fig:results}
\end{figure}
\FloatBarrier

\textbf{Structural property --- variable graph and action dimensions.}
Equations~\eqref{eq:movement-update}--\eqref{eq:lane-update} use shared parameter shapes and return one scalar for every movement presented to them.
Every junction may independently provide \(A_j\) with its own numbers of rows and columns; phase construction is outside the GNN.
This establishes variable graph and action dimensions by construction.
Execution on graphs ranging from 41 to 84 controlled junctions and phase-count ranges of 2--12 validates that the implementation supports the stated structural range.

All learned--baseline comparisons below inherit the decision-interval confound described in Section~3: they compare the recorded implementations, not learning at a matched control frequency.

\textbf{RQ1 --- transfer within the synthetic generator family.}
On the unseen-size \(6\times6\) condition, sampled control had higher throughput and completion than max pressure at all three demands; completion differences were \(3.5,6.6,10.6\) percentage points.
Full means and separate trained-policy and traffic-seed dispersion summaries are reported in Appendix~B.
Orientation-swapped rectangular grids produced closely matched throughput and completion.
These findings are consistent with size and aspect-ratio reuse within the matched generator.

\textbf{RQ2 --- distribution shift.}
The full-coverage-trained policy's throughput and completion deteriorated relative to max pressure at \(50\%\) and \(25\%\) signal coverage, despite the graph remaining structurally executable.
In the separate \(4\times4\) study, sampled throughput differences from max pressure ranged from \(+5.7\) to \(-108.0\) vehicles/hour across coverage conditions, with uncertainty especially wide at \(100\%\); Appendix~B gives the paired intervals.
The result distinguishes architectural compatibility from robustness and is consistent with sensitivity to the controller-distribution shift.

\textbf{RQ3 --- city feasibility.}
The city evaluation reports one policy-training run rather than an estimate across independent training runs.
The policy executed on Karlsruhe, Mannheim, Heidelberg, and Freiburg, all represented in rollout training, and on rollout-held-out Stuttgart.
Stuttgart therefore provides the only city-level test of generalization to a network absent from rollout training in this experiment; the other four results primarily demonstrate shared-parameter execution across heterogeneous training domains.
Outcomes varied: sampled control had higher throughput and completion than every baseline in Karlsruhe and Stuttgart, trailed queue in Mannheim, was close to fixed time in Heidelberg, and traded higher throughput and completion for higher wait density relative to fixed time in Freiburg.
Figure~\ref{fig:results}b uses the strongest non-learned throughput baseline separately for each city.

\section{Limitations and conclusion}

\textbf{Limitations.}
The experiments evaluate the complete implementation and do not isolate individual architectural choices.
They therefore do not determine the separate contributions of typed relations, two message-passing blocks, or sum-based movement-to-phase aggregation; the latter also introduces an unevaluated phase-size bias.
Because only maximal compatible movement sets are enumerated, a compatible strict subset is unavailable whenever another compatible atomic group can be added.
This is a deliberate restriction of the selectable policy space and may prevent a controller from withholding green from an otherwise compatible movement.
Phase enumeration was inexpensive for the evaluated junctions, which produced at most 12 phases, but maximal-compatible-set enumeration can scale poorly for substantially more complex conflict graphs.

The evaluated networks belong to bounded synthetic and city simulation families rather than arbitrary topology.
The coverage study shows sensitivity to the controller distribution, and the city study's one policy-training seed and three traffic seeds per city do not estimate variation across independent training runs.

The generated phase sets inherit the assumptions and fidelity of the SUMO network and implemented conflict rules; they are action sets accepted by the implemented construction, not formally verified real-world signal plans.
Sampled execution outperformed greedy execution in the city study, so the stochastic results do not establish an equally effective deterministic controller.

\textbf{Conclusion.}
The implemented interface separates reusable learned scoring from local signal structure: a shared typed GNN emits one scalar per movement, while deterministic incidence matrices translate those scores into heterogeneous phase spaces.
The experiments add bounded feasibility evidence within synthetic and city simulation families, while the coverage and greedy-execution results show that executable structure alone does not ensure empirical robustness.

\label{lastmainpage}
\clearpage
\bibliographystyle{unsrt}
\bibliography{references}

\clearpage
\appendix
\section{Implementation and scenario details}

\paragraph{Artifact statement.}
Code, committed scenarios, and instructions are available at
\href{https://github.com/BertilBraun/GNN-Traffic-Signal-Control/tree/ea47985}{repository snapshot \texttt{ea47985}}; \texttt{uv.lock} resolves libsumo and sumolib 1.27.1.
The reported protocols use \path{configs/training/grid_shape_generalization_mixed_2hop_gate_30.yaml},
plus the following files under \path{configs/training/}:
\begin{itemize}
  \item \path{grid_signal_coverage_6x6_evaluation.yaml};
  \item \path{grid_coverage_generalization_4x4_train50_2hop_30.yaml} and \path{grid_coverage_generalization_4x4_evaluation.yaml};
  \item \path{city_stuttgart_visible_validation_local_reward_2hop_60.yaml}.
\end{itemize}
Evaluation seeds are recorded in those configurations and the linked result documentation.
\path{scripts/analyze_grid_generalization_results.py} regenerates grid summaries, and \path{scripts/plot_technical_report_results.py} regenerates Figure~\ref{fig:results} from saved CSVs.
The reported trained checkpoints and raw evaluation summaries are preserved in a separate experiment archive, are not part of the Git repository, and are available from the author.

\begin{table}[H]
  \centering
  \caption{Anchored training and control settings.}
  \begin{tabular}{@{}lr@{}}
    \toprule
    Setting & Value\\
    \midrule
    Decision interval / yellow & \(5\,\mathrm{s}\) / immediate \(3\,\mathrm{s}\)\\
    Minimum green & one decision\\
    Message-passing blocks / hidden dimension & 2 / 64\\
    Decisions per rollout / PPO epochs & 200 / 4\\
    Entropy coefficient / reward clip & 0.001 / \([-1,1]\)\\
    Progress/discharge/braking/gridlock & \(1/10/10/0.02\)\\
    Global/flow/throughput/switch/teleport weight & 0\\
    Initial occupancy / warm-up & 6--8\% / \(15\,\mathrm{s}\)\\
    \bottomrule
  \end{tabular}
  \label{tab:settings}
\end{table}

\begin{table}[H]
  \centering
  \caption{Structural diversity of the city scenarios. Phase range is the minimum--maximum number of selectable phases per policy-controlled junction.}
  \begin{tabular}{@{}lrrrr@{}}
    \toprule
    City & Controllers & LaneGroups & Movements & Phase Range\\
    \midrule
    Karlsruhe & 41 & 286 & 271 & 2--10\\
    Mannheim & 84 & 708 & 467 & 2--7\\
    Stuttgart & 69 & 519 & 508 & 2--12\\
    Heidelberg & 56 & 408 & 365 & 2--10\\
    Freiburg & 58 & 425 & 416 & 2--11\\
    \bottomrule
  \end{tabular}
  \label{tab:city-structure}
\end{table}

\section{Detailed evaluation}

\begin{table}[H]
  \footnotesize
  \centering
  \caption{Full-coverage \(6 \times 6\) means and distinct SD sources. Learned means average three trained-policy means, each formed over six traffic seeds; their SD is between trained policies. Max-pressure mean and SD are across six traffic seeds. The SD columns are not directly comparable stability estimates.}
  \begin{tabular}{@{}llrrrr@{}}
    \toprule
    Demand & Metric & Learned mean & Training-seed SD & MP mean & Traffic-seed SD\\
    \midrule
    0.6 & Throughput/h & 4344 & 13 & 4168 & 126\\
        & Completion & 87.2\% & 0.3 pp & 83.6\% & 1.3 pp\\
        & Wait density & 0.077 & 0.008 & 0.082 & 0.062\\
    0.7 & Throughput/h & 4795 & 9 & 4429 & 157\\
        & Completion & 84.8\% & 0.1 pp & 78.2\% & 1.8 pp\\
        & Wait density & 0.114 & 0.009 & 0.119 & 0.120\\
    0.8 & Throughput/h & 5174 & 23 & 4521 & 164\\
        & Completion & 82.2\% & 0.5 pp & 71.5\% & 2.5 pp\\
        & Wait density & 0.175 & 0.018 & 0.131 & 0.022\\
    \bottomrule
  \end{tabular}
  \label{tab:grid-detail}
\end{table}

Within each trained policy, traffic-seed SDs ranged from \(121\) to \(168\) vehicles/hour for throughput, \(0.4\) to \(2.0\) percentage points for completion, and \(0.007\) to \(0.024\,\mathrm{s/m}\) for wait density across the three demands.
Thus the small between-policy SDs in Table~\ref{tab:grid-detail} should not be read as episode-level variability.

\clearpage
\begin{table}[H]
  \footnotesize
  \centering
  \caption{Iteration-60 city means \(\pm\) traffic-seed SD over three seeds. The baseline has the highest mean throughput among non-learned policies in that city. Wait density is in seconds per metre; these SDs do not include trained-policy variability.}
  \begin{tabular}{@{}llrrr@{}}
    \toprule
    City & Policy & Throughput/h & Completion & Wait density\\
    \midrule
    Karlsruhe & learned sampled & \(3452\pm314\) & \(93.86\pm0.34\)\% & \(0.075\pm0.034\)\\
               & max pressure & \(3105\pm241\) & \(84.77\pm8.12\)\% & \(0.243\pm0.226\)\\
    Mannheim & learned sampled & \(3200\pm563\) & \(53.78\pm13.46\)\% & \(0.796\pm0.418\)\\
              & queue & \(3394\pm562\) & \(56.78\pm11.96\)\% & \(1.019\pm0.563\)\\
    Stuttgart & learned sampled & \(4216\pm528\) & \(50.93\pm9.13\)\% & \(0.907\pm0.597\)\\
               & fixed time & \(3661\pm154\) & \(43.97\pm1.18\)\% & \(0.741\pm0.227\)\\
    Heidelberg & learned sampled & \(3190\pm127\) & \(77.69\pm1.53\)\% & \(0.181\pm0.055\)\\
                & fixed time & \(3250\pm134\) & \(79.16\pm2.19\)\% & \(0.127\pm0.044\)\\
    Freiburg & learned sampled & \(2670\pm225\) & \(48.39\pm6.77\)\% & \(1.108\pm0.449\)\\
              & fixed time & \(2547\pm35\) & \(45.92\pm2.67\)\% & \(0.934\pm0.242\)\\
    \bottomrule
  \end{tabular}
  \label{tab:city-detail}
\end{table}

\paragraph{Sampled versus greedy.}
For the unweighted mean across cities, greedy execution achieved \(3182.6\) vehicles/hour, \(61.89\%\) completion, and \(0.980\,\mathrm{s/m}\) wait density, compared with \(3345.6\), \(64.93\%\), and \(0.613\,\mathrm{s/m}\) for sampled execution.

\paragraph{Coverage study.}
The \(50\%\)-coverage training study used five training layouts and three held-out traffic seeds on each of five evaluation layouts at \(25\%\), \(50\%\), and \(75\%\) coverage; the \(100\%\) condition used one layout and the same three seeds.
Each paired observational unit is one evaluation layout--traffic-seed combination: \(n=15\) at each partial-coverage condition and \(n=3\) at \(100\%\) coverage.
For sampled execution, PPO \( - \) max-pressure throughput differences at \(25\%\), \(50\%\), \(75\%\), and \(100\%\) coverage were \(+5.7\), \(-28.9\), \(-104.3\), and \(-108.0\) vehicles/hour; completion differences were \(+0.24\), \(-0.66\), \(-2.99\), and \(-3.14\) percentage points.
Paired 95\% Student-\(t\) interval half-widths for throughput were \(52.5,54.1,41.3,345.2\) vehicles/hour, respectively; the \(100\%\) condition consequently has a very wide interval.

\clearpage
\section{Concrete graph example}

\begin{figure}[H]
  \centering
  \includegraphics[width=0.92\linewidth]{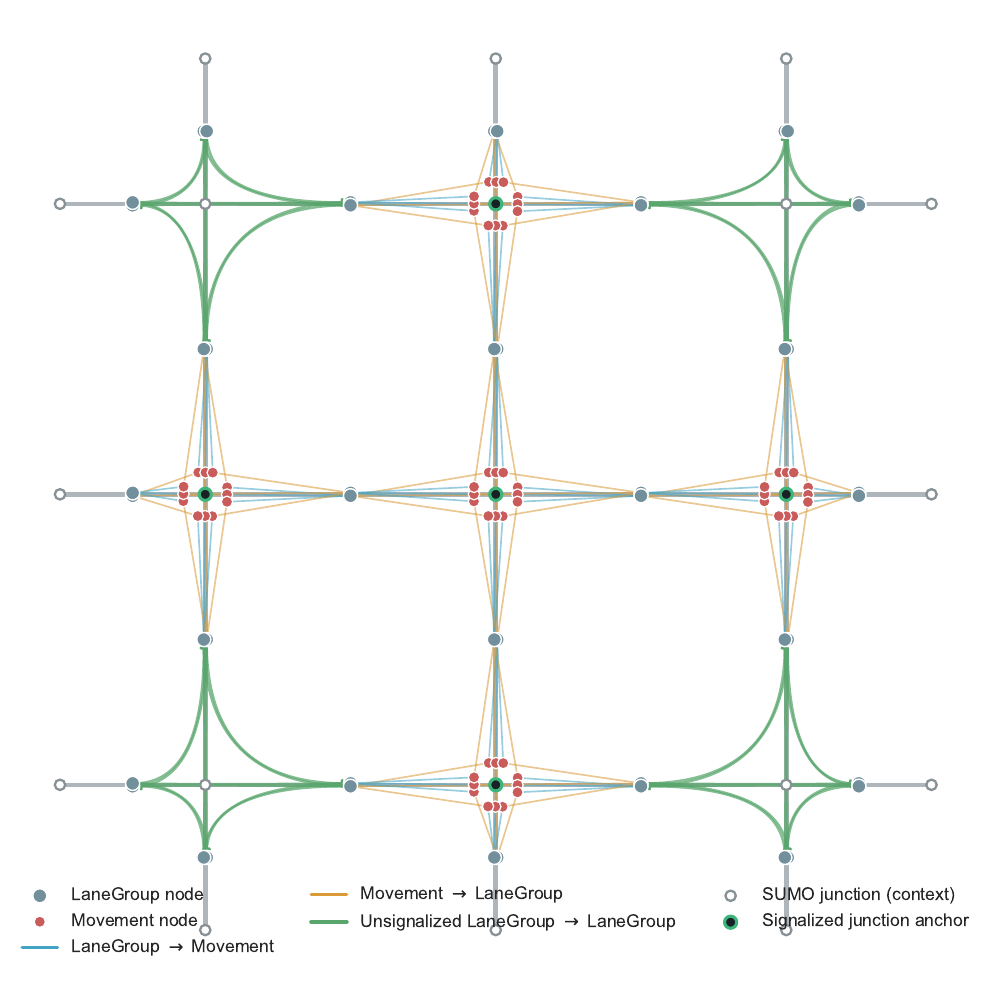}
  \caption{Complete \(3\times3\) synthetic road layout represented as the controller graph. LaneGroup nodes follow directed corridors; Movement nodes encode legal input-to-output paths at signalized junctions. Cyan and amber edges are the typed LaneGroup-to-Movement and Movement-to-LaneGroup relations, while green edges propagate LaneGroup context through unsignalized junctions. Junction markers provide spatial context only and are not GNN nodes. Positions are generated from the same embedded graph data used by the interactive HTML visualization.}
  \label{fig:full-movement-graph}
\end{figure}

\end{document}

%% file: figures/intersection-concepts.tex
\begin{tikzpicture}[
  x=1cm,
  y=1cm,
  panel/.style={
    draw=gray!35,
    rounded corners=1.5mm,
    fill=gray!2,
    minimum width=3.75cm,
    minimum height=3.35cm
  },
  road/.style={fill=gray!18, draw=gray!35},
  movement/.style={->, >=Stealth, line width=2.1pt, color=orange!85!black},
  active/.style={->, >=Stealth, line width=2.2pt, color=teal!80!black},
  inactive/.style={->, >=Stealth, line width=1.2pt, color=gray!55},
  label/.style={font=\small, align=center}
]
  \foreach \offset in {0,4.35,8.7} {
    \begin{scope}[xshift=\offset cm]
      \node[panel] at (0,0) {};
      \path[road] (-1.55,-0.34) rectangle (1.55,0.34);
      \path[road] (-0.34,-1.35) rectangle (0.34,1.35);
      \draw[white, line width=0.6pt, dashed] (-1.5,0) -- (1.5,0);
      \draw[white, line width=0.6pt, dashed] (0,-1.3) -- (0,1.3);
    \end{scope}
  }

  \begin{scope}
    \path[fill=blue!16] (-0.30,-1.30) rectangle (0.30,-0.15);
    \path[fill=blue!16] (0.15,-0.30) rectangle (1.50,0.30);
    \draw[movement] (0,-0.95) .. controls (0,-0.35) and (0.35,0) .. (1.02,0);
    \node[label, text=blue!70!black] at (-0.95,-1.05) {incoming\\corridor};
    \node[label, text=blue!70!black] at (1.05,0.73) {outgoing\\corridor};
    \node[font=\small\bfseries] at (0,1.62) {(a) One movement};
  \end{scope}

  \begin{scope}[xshift=4.35cm]
    \draw[movement] (-0.18,-1.05) -- (-0.18,1.02);
    \draw[movement] (1.06,-0.16) -- (-1.02,-0.16);
    \draw[movement] (0.18,1.05) .. controls (0.18,0.35) and (0.35,0.18) .. (1.05,0.18);
    \node[label, fill=white, inner sep=1pt] at (-0.60,0.62) {$s_1$};
    \node[label, fill=white, inner sep=1pt] at (0.58,-0.52) {$s_2$};
    \node[label, fill=white, inner sep=1pt] at (0.70,0.63) {$s_3$};
    \node[font=\small\bfseries] at (0,1.62) {(b) Score each movement};
  \end{scope}

  \begin{scope}[xshift=8.7cm]
    \draw[active] (-0.18,-1.05) -- (-0.18,1.02);
    \draw[active] (0.18,1.05) -- (0.18,-1.02);
    \draw[inactive] (-1.06,0.18) -- (1.02,0.18);
    \draw[inactive] (1.06,-0.18) -- (-1.02,-0.18);
    \node[label, text=teal!80!black, fill=white, inner sep=1pt] at (0.78,0.74)
      {green\\together};
    \node[label, text=gray!80, fill=white, inner sep=1pt] at (-0.78,-0.68)
      {held\\red};
    \node[font=\small\bfseries] at (0,1.62) {(c) Select one phase};
  \end{scope}
\end{tikzpicture}

%% file: figures/representation.tex
\begin{tikzpicture}[
  x=1cm,
  y=1cm,
  lane/.style={circle, draw=blue!65!black, fill=blue!12, minimum size=9mm, inner sep=0pt},
  movement/.style={circle, draw=orange!75!black, fill=orange!22, minimum size=8mm, inner sep=0pt},
  relation/.style={<->, >=Stealth, line width=0.8pt, color=gray!75},
  influence/.style={->, >=Stealth, line width=1.5pt, color=teal!75!black},
  block/.style={
    draw=gray!45,
    rounded corners=1.5mm,
    fill=gray!5,
    text width=4.2cm,
    minimum height=10mm,
    font=\small,
    align=center
  },
  note/.style={font=\small, align=center}
]
  \node[lane] (l0) at (0,0) {$L_0$};
  \node[movement] (m0) at (1.8,0) {$M_0$};
  \node[lane] (l1) at (3.6,0) {$L_1$};
  \node[movement] (m1) at (5.4,0) {$M_1$};
  \node[lane] (l2) at (7.2,0) {$L_2$};

  \draw[relation] (l0) -- (m0);
  \draw[relation] (m0) -- (l1);
  \draw[relation] (l1) -- (m1);
  \draw[relation] (m1) -- (l2);

  \draw[influence, bend left=31] (m0.north) to
    node[above, note] {neighbour context reaches $M_1$ in block 2}
    (m1.north);

  \node[block] (block1) at (2.15,-1.25)
    {\textbf{Block 1}\\LaneGroups $\rightarrow$ Movements\\$\rightarrow$ LaneGroups};
  \node[block] (block2) at (7.65,-1.25)
    {\textbf{Block 2}\\repeat using updated context};
  \draw[influence] (block1.east) -- (block2.west);
\end{tikzpicture}

%% file: figures/phase-interface.tex
\begin{tikzpicture}[
  x=1cm,
  y=1cm,
  box/.style={draw=gray!55, rounded corners=1.5mm, fill=gray!4, minimum height=9mm, align=center, font=\small},
  arrow/.style={->, >=Stealth, line width=0.9pt, color=gray!75},
  learned/.style={draw=teal!70!black, fill=teal!8},
  fixed/.style={draw=blue!60!black, fill=blue!7}
]
  \node[box] (links) at (0,1.2) {SUMO controlled\\links and conflicts};
  \node[box] (groups) at (3.0,1.2) {atomic activation\\groups};
  \node[box] (cliques) at (6.0,1.2) {maximal compatible\\sets};
  \node[box, fixed] (incidence) at (9.0,1.2) {local incidence\\matrix $A_j$};
  \draw[arrow] (links) -- (groups);
  \draw[arrow] (groups) -- (cliques);
  \draw[arrow] (cliques) -- (incidence);
  \node[font=\small\itshape, text=gray!70!black] at (4.5,2.0) {offline, deterministic construction};

  \node[box, learned] (scores) at (0,-0.5) {GNN Movement\\scores $\mathbf{s}_j$};
  \node[font=\Large] at (2.15,-0.5) {$\times$};
  \node[box, fixed] (matrixcopy) at (3.4,-0.5) {$A_j$};
  \node[font=\Large] at (4.7,-0.5) {$=$};
  \node[box] (logits) at (6.0,-0.5) {phase logits\\$\boldsymbol{\ell}_j=A_j\mathbf{s}_j$};
  \node[box] (mask) at (8.55,-0.5) {availability\\mask};
  \node[box, learned] (action) at (11.0,-0.5) {sample one\\available phase};
  \draw[arrow] (logits) -- (mask);
  \draw[arrow] (mask) -- (action);
  \draw[arrow, dashed] (incidence.south) -- (matrixcopy.north);
  \node[font=\small\itshape, text=gray!70!black] at (8.45,-1.35) {online, once per five-second decision};
\end{tikzpicture}

%% file: references.bib
@article{varaiya2013maxpressure,
  author = {Varaiya, Pravin},
  title = {Max Pressure Control of a Network of Signalized Intersections},
  journal = {Transportation Research Part C: Emerging Technologies},
  volume = {36},
  pages = {177--195},
  year = {2013},
  doi = {10.1016/j.trc.2013.08.014}
}

@inproceedings{wei2019presslight,
  author = {Wei, Hua and Chen, Chacha and Zheng, Guanjie and Wu, Kan and Gayah, Vikash and Xu, Kai and Li, Zhenhui},
  title = {{PressLight}: Learning Max Pressure Control to Coordinate Traffic Signals in Arterial Network},
  booktitle = {Proceedings of the 25th ACM SIGKDD International Conference on Knowledge Discovery and Data Mining},
  pages = {1290--1298},
  year = {2019},
  doi = {10.1145/3292500.3330949}
}

@inproceedings{zheng2019frap,
  author = {Zheng, Guanjie and Xiong, Yuanhao and Zang, Xinshi and Feng, Jie and Wei, Hua and Zhang, Huichu and Li, Yong and Xu, Kai and Li, Zhenhui},
  title = {Learning Phase Competition for Traffic Signal Control},
  booktitle = {Proceedings of the 28th ACM International Conference on Information and Knowledge Management},
  pages = {1963--1972},
  year = {2019},
  doi = {10.1145/3357384.3357900}
}

@article{schmidt2024transferlight,
  author = {Schmidt, Johann and Dreyer, Frank and Hashimi, Sayed Abid and Stober, Sebastian},
  title = {{TransferLight}: Zero-Shot Traffic Signal Control on any Road-Network},
  journal = {arXiv preprint arXiv:2412.09719},
  year = {2024},
  doi = {10.48550/arXiv.2412.09719}
}

@article{schulman2017ppo,
  author = {Schulman, John and Wolski, Filip and Dhariwal, Prafulla and Radford, Alec and Klimov, Oleg},
  title = {Proximal Policy Optimization Algorithms},
  journal = {arXiv preprint arXiv:1707.06347},
  year = {2017},
  doi = {10.48550/arXiv.1707.06347}
}

@inproceedings{alvarezlopez2018sumo,
  author = {Alvarez Lopez, Pablo and Behrisch, Michael and Bieker-Walz, Laura and Erdmann, Jakob and Fl{\"o}tter{\"o}d, Yun-Pang and Hilbrich, Robert and L{\"u}cken, Leonhard and Rummel, Johannes and Wagner, Peter and Wie{\ss}ner, Evamarie},
  title = {Microscopic Traffic Simulation using {SUMO}},
  booktitle = {2018 21st International Conference on Intelligent Transportation Systems},
  pages = {2575--2582},
  year = {2018},
  doi = {10.1109/ITSC.2018.8569938}
}

@article{bron1973cliques,
  author = {Bron, Coen and Kerbosch, Joep},
  title = {Algorithm 457: Finding All Cliques of an Undirected Graph},
  journal = {Communications of the ACM},
  volume = {16},
  number = {9},
  pages = {575--577},
  year = {1973},
  doi = {10.1145/362342.362367}
}
